\documentclass[letterpaper, 10 pt, conference]{ieeeconf}  % Comment this line out if you need a4paper
\usepackage{graphicx}      % for \includegraphics
\usepackage{subcaption}    % for \begin{subfigure} … \end{subfigure}
\graphicspath{{figs/}}     % tell LaTeX to look in “figs/” for images

\IEEEoverridecommandlockouts                              % This command is only needed if 
\makeatletter
\let\NAT@parse\undefined
\makeatother

\usepackage{hyperref}
\usepackage[utf8]{inputenc}
\usepackage{authblk}
\usepackage{bbm}
\usepackage{amsmath}
\usepackage{algorithm}
\usepackage[noend]{algpseudocode}

\usepackage{marginnote}
\usepackage{amsfonts}
\usepackage{amsmath}
\usepackage{graphicx}
\usepackage[usenames,dvipsnames,table,xcdraw]{xcolor}
\usepackage{xspace}
\usepackage{caption}
\usepackage[table]{xcolor}
\usepackage{multirow}

\usepackage[usenames,dvipsnames,table,xcdraw]{xcolor}
\usepackage{colortbl}
\usepackage{multirow}
\usepackage{nicematrix}
\usepackage{makecell}

\definecolor{lightblue}{rgb}{0.68, 0.85, 0.9}
\definecolor{lighterblue}{rgb}{0.80, 0.92, 0.95}
\definecolor{lightergray}{rgb}{0.90, 0.90, 0.90}
\definecolor{lighttan}{rgb}{0.82, 0.71, 0.55}
\definecolor{lightertan}{rgb}{0.94, 0.87, 0.80}
\definecolor{lightgreen}{rgb}{0.56, 0.93, 0.56}
\definecolor{lightergreen}{rgb}{0.74, 0.99, 0.79}
\definecolor{lightorange}{rgb}{1.00, 0.78, 0.49}
\definecolor{lighterorange}{rgb}{1.00, 0.88, 0.70}
\definecolor{lightred}{rgb}{1.0, 0.5, 0.5}
\definecolor{lighterred}{rgb}{1.0, 0.6, 0.6}

\usepackage[symbol]{footmisc}

\newcommand{\Term}[1]{\textsf{#1}}

\usepackage{subcaption}

\usepackage[nameinlink]{cleveref}

\definecolor{almostblack}{rgb}{0, 0, 0.3}

\newcommand{\FK}{\mathtt{FK}}

\hypersetup{%
      breaklinks,%
      colorlinks=true,%
      urlcolor=[rgb]{0.25,0.0,0.0},%
      linkcolor=[rgb]{0.5,0.0,0.0},%
      citecolor=[rgb]{0,0.2,0.445},%
      filecolor=[rgb]{0,0,0.4},
      anchorcolor=[rgb]={0.0,0.1,0.2}%
}

\newcommand{\cspace}{\ensuremath{\mathcal{C}_{space}}}

\usepackage{todonotes}

\newcommand{\R}{\mathbb{R}}

\newcommand{\1}{\mathbbm{1}}
\newcommand{\eps}{\varepsilon}
\newcommand{\boundary}{\partial}

\newcommand{\Set}[2]{\left\{ #1 \;\middle\vert\; #2 \right\}}

\newcommand{\set}[1]{\left\{ {#1} \right\}}

\newcommand{\RGG}{\Term{RGG}\xspace}
\newcommand{\DOF}{\Term{DOF}\xspace}
\newcommand{\RRT}{\Term{RRT}\xspace}

\newcommand{\PRM}{\Term{PRM}\xspace}
\newcommand{\SBMP}{\Term{SBMP}\xspace}

\newcommand{\AABB}{\Term{AABB}\xspace}
\newcommand{\OBB}{\Term{OBB}\xspace}
\newcommand{\LazyPRM}{\Term{LazyPRM}\xspace}

\newcommand{\DRM}{\Term{DRM}\xspace}
\newcommand{\cfg}{\mathtt{cfg}\xspace}
\newcommand{\dist}{\mathtt{dist}\xspace}

\newcommand{\etal}{\textit{et~al.}\xspace}

\newcommand{\si}[1]{#1}

\newcommand{\HLink}[2]{\hyperref[#2]{#1~\ref*{#2}}}

\title{\LARGE \bf
Quick Heuristic Validation of Edges in Dynamic Roadmap Graphs
}

\author{Yulie Arad$^{1}$ and Stav Ashur$^{1}$ and Nancy M. Amato$^{1}$% 
\thanks{$^{1}$ Siebel School of Computing and Data Science, University of Illinois, 201 N. Goodwin Avenue, Urbana, IL 61801, USA}%
}

\begin{document}

\maketitle
\thispagestyle{empty}
\pagestyle{empty}

\begin{abstract}
    % Roadmap graphs are widely used in robot motion planning, and offer an effective solution for multi-query scenarios where a robot is required to perform many motion planning tasks in the same environment. Non-static environments in which some objects or other agents move require dynamic roadmaps, capable of quickly updating the validity status of their nodes and edges as changes occur. The main challenge encountered by these data structures is edge verification, as edges represent swept volumes of possibly complex robots.

    In this paper we tackle the problem of adjusting roadmap graphs for robot motion planning to non-static environments. We introduce the ``Red-Green-Gray'' paradigm, a modification of the SPITE method \cite{almmmhpa-spitme-24}, capable of classifying the validity status of nodes and edges using cheap heuristic checks, allowing fast semi-lazy roadmap updates. Given a roadmap, we use simple computational geometry methods to approximate the swept volumes of robots and perform lazy collision checks, and label a subset of the edges as invalid (red), valid (green), or unknown (gray).

    We present preliminary experimental results comparing our method to the well-established technique of Leven and Hutchinson \cite{lh-frtppce-02}, and showing increased accuracy as well as the ability to correctly label edges as invalid while maintaining comparable update runtimes. 

 \end{abstract}

\section{Introduction}
A motion planning problem consists of an environment (workspace) $W$, a Robot $r$ and start and goal configurations, $s$ and $t$, where the aim is to find a valid $(s,t)$-path for $r$. Many motion planning algorithms create geometric graphs, i.e. roadmaps, in the robot \emph{configuration space} (c-space), the implicit space of all configurations of the robot in the environment. Nodes of the graph correspond to individual configurations of the robot, and edges to continuous motions (sequences of configurations).  

In order for a roadmap edge to be used in a path it must be validated, usually by checking a finite set of intermediate points along it for collisions with obstacles and, when applicable, self collisions (e.g., in articulated robots). The computational complexity of collision detection increases dramatically with the number of degrees of freedom (\DOF) of the robot, and increases further with the complexity of the environment and the robot. Since collision checking of a dense set of intermediates is usually required to ensure a valid path, edge validation operations become the computational bottleneck in many motion planning algorithms \cite{ksh-cdnnscbsbmp-16}.

Some roadmap graphs, e.g. ones created by the Probabilistic Roadmap (\PRM) algorithm \cite{kslo-prpp-96}, are data structures designed for multiple motion planning queries. The purpose of the roadmap in this case is to capture the structure and connectivity of the c-space associated with the problem, and reduce motion planning queries to graph pathfinding queries by connecting the start and goal configurations $s$ and $t$ to the graph with edges, and returning an $(s,t)$-path on the graph. Such data structures are especially suitable for use when the environment is fixed, say, a single warehouse or factory, in which a robot is routinely required to navigate.   

\begin{figure}[t!]
    \centering 
    \subfloat[]{
       \includegraphics[width=0.75\linewidth,page=6]{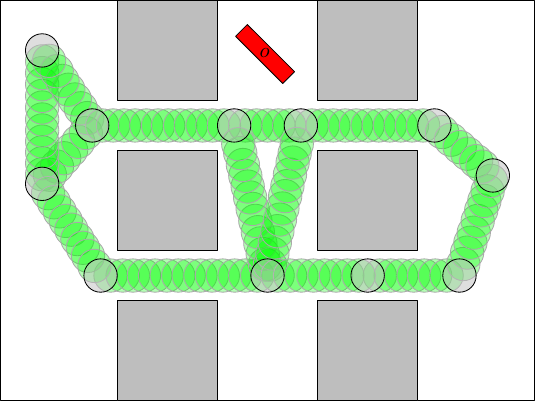}
       \label{fig:dynamic:roadmap:before}}
       \vfill
    \subfloat[]{
        \includegraphics[width=0.75\linewidth,page=5]{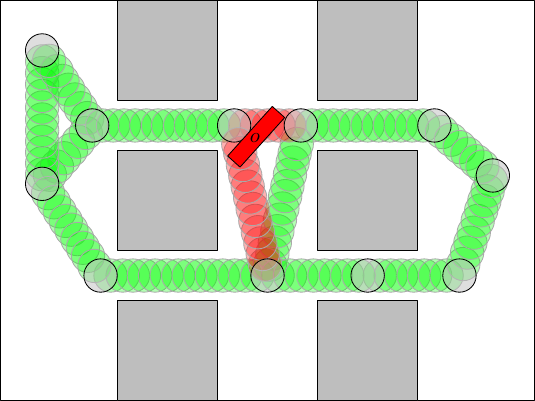}
       \label{fig:dynamic:roadmap:after}}
    \caption{An illustration of a roadmap for a 2D translational disk robot in a 2D environment. The edges, shown as sequences of robot configuration along line segments, may become invalid as obstacles change their positions. When the obstacle, labeled $o$, moves from its first position, shown in Figure \ref{fig:dynamic:roadmap:before}, to another position, shown in Figure \ref{fig:dynamic:roadmap:after}, two edges become invalid (red).}
    \label{fig:dynamic:roadmap}
\end{figure}

In this paper, we consider the problem of adjusting such roadmaps for use in a dynamic environment where obstacles change their positions or in which other agents are moving through. See Figure \ref{fig:dynamic:roadmap} for an illustration. The ability to efficiently handle dynamic environments is crucial for practical use in modern-day robotics, as even in the mostly static environments mentioned above a certain amount of movement still occurs, whether it is a door opening or closing, objects being moved around, or humans performing some task. In such environments, re-running a single query motion planning algorithm can be quite costly and inefficient, given that changes are mostly local, and much of the environment, walls, shelves, heavy machinery etc., movements are rare. 

\subsection*{Contribution}

We present a method to maintain a roadmap in the c-space under configuration space updates. As obstacles move around, some portion of the graph might remain valid, while others might be invalid, or ``suspicious'' -- requiring revalidation. By effectively maintaining this graph, future tasks can avoid expensive fine-grained computation. We build upon the SPITE method that bounds the swept volumes corresponding to the graph's nodes and edges using \emph{cigars} (also known as line-swept spheres or capped cylinders) in order to quickly determine if, say, an edge is collision free.

\paragraph*{The starting point} 
The precursor to \RGG, SPITE, could only find valid nodes and edges by using outer-approximations of swept volumes. We introduce the ``Red-Green-Gray'' (\RGG) paradigm that uses both outer- and inner-approximations to quickly and accurately label a portion of the roadmap as invalid (red), valid (green), or of unknown validity (gray). We replace SPITE's cigars with oriented bounding boxes (\OBB) in order to over-approximate the different volumes and label graph components as valid, and use splines based on sphere approximations for polyhedrons to get under-approximations and label components as invalid. Edges that cannot be easily classified as valid or invalid are labeled ``unknown'', and can be validated during query evaluation, trading-off runtime for more optimal paths when using the \RGG method to answer motion planning queries.

We demonstrate our method's improved accuracy in determining the validities of edges compared to the well-known grid-based dynamic roadmap of Leven and Hutchinson, which is, to the best of our knowledge, the only tool created for adjusting general c-space graphs for modifiable environments. We show that our method achieves higher levels of accuracy when determining validities of nodes and edges, while maintaining comparable update runtimes.

\subsection*{Organization}
%\paragraph*{Paper layout}
In Section \ref{sec:related:work} we review some of the previous work on the topics discussed here. In Section \ref{sec:preliminaries} we introduce definitions and notations required in the following  Section \ref{sec:method} in which we describe the approximation algorithms that make up our method and the way they are combined to create the \RGG scheme. In Section \ref{sec:experiment} we present the experiments conducted to evaluate our contribution, followed by a discussion of the results and future work in Section \ref{sec:discussion}.

\section{Related Work}
\label{sec:related:work}
Roadmap graphs were popularized by Kavraki \etal for robot motion planning by the introduction of the Probabilistic Roadmap algorithm (\PRM) \cite{kslo-prpp-96}, which gave rise to a host of sampling-based motion planning (\SBMP) algorithms. See \SBMP ~\cite{ock-sbmpcr-23} and references therein for a comprehensive review of \SBMP. Some \SBMP algorithms, like \PRM, construct a roadmap meant to be used as a data structure capable of efficiently answering multiple motion planning queries for a single robot in a fixed environment. Others, like \LazyPRM \cite{bk-ppulp-00} are focused on answering a single query, making them useful building blocks for many algorithms used in dynamic settings.   

\LazyPRM creates a roadmap in the c-space using random sampling, but unlike \PRM it does not check the validity of the edges unless they are on a path considered as an output, and thus avoids expensive collision checking in some areas of c-space. Some variants of \LazyPRM also don't validate roadmap nodes \cite{bk-ararpp-01,bk-ppulp-00}, and some implementations approximate the validity of nodes and edges before full validation is done \cite{sta-gfpmp-03}. Jaillet and Sim\`{e}on~\cite{js-apbmpdce-04} used a static roadmap for static obstacles, and postponed collision checking edges against obstacles whose position had changed to query time with an adaptation of \LazyPRM.

Dynamic roadmap graphs that can be updated as obstacles change position have been studied by Leven and Hutchinson \cite{lh-frtppce-02}, whose algorithm partitions the environment using a grid, and maps each grid cell to the nodes and edges that intersect it, that is that contain configurations in which the robot intersects the cell. The obstacle's axis-aligned bounding box (\AABB) can then be checked for intersection with cells in order to quickly find possible interference between the roadmap and the obstacle. This dynamic roadmap, commonly known as \DRM, was later implemented and tested by Kallmann and Matari\`{c} \cite{km-mpdr-04}, who showed that it is more efficient in answering motion planning queries in a modified environment than \RRT \cite{l-rrtntpp-1998}. Yang \etal \cite{ymilv-hdrmrcdrrtmpcs-17} created a hierarchical \DRM for articulated robots, which uses the hierarchy induced by the ordering of the bodies composing the robot to efficiently update sets of nodes and edges in a grid roadmap. 

The SPITE method \cite{almmmhpa-spitme-24}, which this paper extends, focused on approximating swept volumes created by motions of the robot. SPITE uses geometric shapes called cigars (also known as line-swept spheres or capped cylinders) to approximate the swept volumes of the robot, and rule out many nodes and edges whose validity status can be determined without exact collision. This approach was shown to result in improved motion planning runtimes when compared to \DRM, \RRT, and \LazyPRM. While SPITE and \RGG use rudimentary tools, a large body of work is dedicated to the computation of swept volumes \cite{amybj-swfpa-06} and collision detection \cite{lmk-cpq-17}, which are fundamental for many fields of engineering. Larsen \etal \cite{lglm-pqp-99,lglm-fdqrssv-00} have studied the intersection of the two, and gave collision detection and proximity queries for swept volumes of simple shapes. 

\section{Preliminaries}
\label{sec:preliminaries}

Let $R = (V,E)$ be a roadmap graph in the implicit c-space $\cspace$ defined by a robot $r$ in a 3D environment (workspace) $W$ containing a set of obstacles $O$. We refer to $V\cup E$, the set of nodes and edges of $R$, as the \emph{components} of the roadmap. Note that this reflects the applicability of our methods for both nodes and edges, since, for our purposes, nodes can be viewed as degenerate edges.

While roadmap edges are continuous curves in c-space, most algorithms approximate them into discrete sequences of configurations. We assume that every edge $e$ is composed of $\frac{||e||}{\eps}$ configurations, where $||e||$ is the length of the edge, and $\eps > 0$ is the resolution parameter, and denote the set $\cfg(e)$.

The \emph{forward kinematics} function
\begin{equation*}
    \FK:\cspace \longrightarrow P(\R^3)
\end{equation*}
maps sequences of \DOF{}s (c-space coordinates) to configurations of $r$ in $W$. With some abuse of notation, we also use $\FK$ to mean its natural extension for sets $e\subseteq \cspace$,

\begin{align*}
        \FK:(e) = \bigcup_{c\in \cfg(e)} \FK(c),
\end{align*}

and also the mapping of coordinates of objects from $O$ to their 3D volumes. We refer to the output of the forward kinematics function as a \emph{swept volume} (even if its input is a single point from a c-space).

An outer-approximation (inner-approximation) of a compact set $C\subseteq \R^3$ is a superset (subset) of $C$, that is, a set $C'$  such that $C\subseteq C'$ ($C'\subseteq C$).

%%%%%%%%%%%%%%%%%%%%%%%%%%%%%%%%%%%

\section{Method}
\label{sec:method}

The idea behind the new \RGG paradigm is simple. We use two types of heuristic subroutines that quickly determine the validity of a subset of the graph, thus estimating the roadmap's state as accurately as possible. Both subroutines use an approximation of the swept volumes of roadmap components and obstacles, which are created during preprocessing and stored in an \AABB-trees for quick intersection checks. One subroutine checks for intersections (or lack thereof) between outer-approximations of obstacles and swept volumes of the robot, and the other performs similar checks for inner-approximations.

\begin{figure}[t!]
    \centering 
    \subfloat[Outer- and inner-approximations in a 2D scenario with a fixed-base manipulator]{
       \includegraphics[width=0.7\linewidth,page=5]{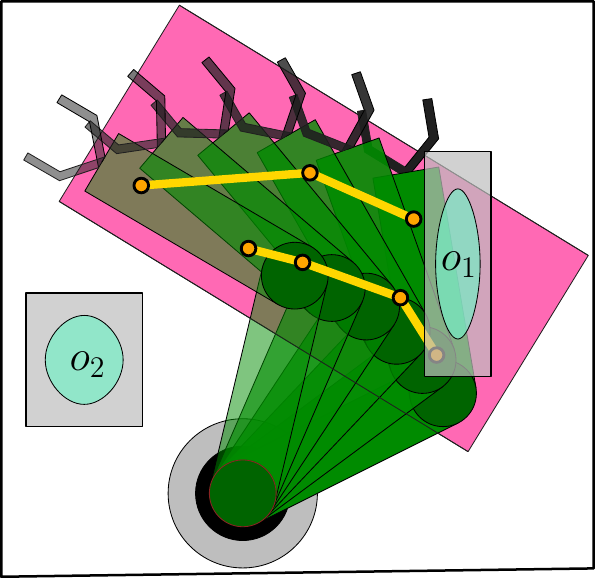}
       \label{fig:rgg:example:manip}}
       \vfill
    \subfloat[Outer- and inner-approximations in a 2D scenario with a mobile robot]{
        \includegraphics[width=0.7\linewidth,page=4]{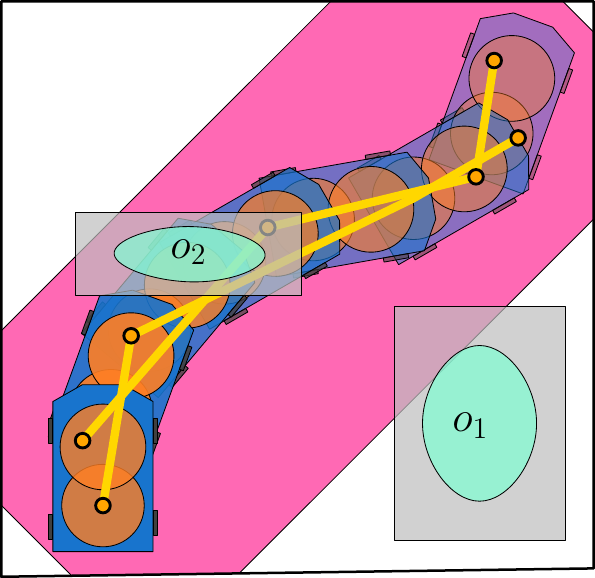}
       \label{fig:rgg:example:mobile}}
    \caption{Two examples of outer- and inner-approximations of a swept volume created by motion of a robot (corresponding to a roadmap edge). One link of the manipulator and the entire mobile robot are over-approximated by an oriented bounding box (pink) and under-approximated by two splines (yellow), and obstacles are over-approximated by an \AABB (gray) and under-approximated by an oval shape (light blue). In Figure \ref{fig:rgg:example:manip} $o_1$ gives rise to a gray edge (edge is invalid), and $o_2$ gives rise to a green edge. In Figure \ref{fig:rgg:example:mobile} $o_1$ gives rise to a gray edge (edge is valid), and $o_2$ gives rise to a red edge.}
    \label{fig:rgg:example}
\end{figure}

An intersection between two inner-approximations implies that a component of the graph is invalid (i.e., red), while non-intersecting outer-approximations implies it is valid (i.e., green). If, say, an edge's and an obstacle's outer-approximations intersect, but their inner-approximations do not, we label the edge as having unknown validity (gray). See Figure \ref{fig:rgg:example} for examples. First, this laziness avoids unnecessary fine-grained (and computationally expensive) collision checking if the obstacle continues its motion or another obstacle moves in to invalidate the edge. Furthermore, if and when a motion planning query is given, we do not spend excessive resources on edges that are irrelevant.
%serve no purpose in answering it.

\paragraph*{Choosing approximation methods} Our choices of approximation methods were driven by two different optimization objectives. Let $x \in V \cup E$ be a component of $R$ with outer-approximation $x^+$ and inner-approximation $x^-$, and let $o$ be an obstacle in $W$ with outer-approximation $o^+$ and inner-approximation $o^-$. 

For every such $x$ and $o$, we want to minimize the time spent on the collision detection/proximity queries $(x^+,o^+)$ and $(x^-,o^-)$, while minimizing
\begin{gather*}
    |\1[x^+ \cap o^+ \neq \emptyset] - \1[\FK(x) \cap \FK(o) \neq \emptyset]|%
    \\%
    \text{and }%
    \\%
    |\1[x^- \cap o^- \neq \emptyset] - \1[\FK(x) \cap \FK(o) \neq \emptyset]|,
\end{gather*}
where $\1[X]$ is the indicator function of event $X$. These latter terms are formal representations of the objective of capturing intersections of the underlying volumes using their approximations.

Taking these objectives into account, we chose \OBB{}s and splines for outer- and inner-approximations of swept volumes of graph components (respectively), and \AABB{}s and sets of spheres for outer- and inner-approximations of obstacles (respectively). We now describe the computation of the approximations in greater detail.

% \paragraph*{TODO} add a description of the RGG method, how we use the over and under approximations for collision checking - should emphasize (here and in the paper in general) that the important thing is that we're lazily collision checking entire edges, and how the RGG framework is used in dynamic roadmaps.  

% Edges are classified in to three groups, "red", "green" and "gray". Quick geometric checks are performed in order to create these classifications. These checks, over- and under-approximation measurements, will be explained further later. 

% An edge is "red" if it is determined to be invalid for this particular environment. That is, there exists some obstacle interfering with the robot as it moves along the edge. The under-approximation method we present is able to find edges that should be marked as "red". 

% Similarly, an edge is marked "green" if the over-approximation determines that no obstacles intersect the edge. A "gray" edge is one that is found to be inconclusive by both the under and over-approximation. "Gray" edges need to be validated using collision detection (side note, don't know how to say this). 

% add implementation details, one AABB tree of over and under and check this and this 
\subsection{Oriented bounding boxes}
Since swept volumes created by motions of robots can create elongated shapes, shapes like cigars and \OBB{}s provide a low complexity reasonably accurate approximation.

\paragraph*{Computing \OBB{}s} For an edge $e$, let $I = \bigcup_{c\in e} V(\FK(c))$, where $V(\cdot)$ is the set of vertices of the input polyhedral shape. Simply put, $I$ is the union of vertices of the robot over $\cfg(e)$. We create an outer-approximation of $e$ by using an algorithm of Barequet and Har-Peled \cite{bhp-eamvbbpstd-01} that returns an approximation of the minimum volume oriented bounding box (\OBB) of a 3D point cloud. Specifically, we use Har-Peled's open source implementation of the algorithm \cite{hp-sccadps-00}. We do not elaborate on this algorithm further due to space constraints.

For articulated robots composed of $k$ bodies, the point cloud created by each body is approximated separately and mapped to the appropriate graph component.

% In order to over-approximate an edge, we create an oriented bounding box (OBB) for each body that completely encapsulates the body as it moves from one configuration to another, including all intermediates. 

% For each body, for each edge $(s,t)$, consider the set of vertices of the body in the starting configuration $s$. Let this set be $V_s$. Let $V_i$ be the set of vertices of the body in configuration $i$, where $i$ is the $i$-th intermediate and let $V_t$ be the set of vertices of the body in the final configuration $t$. Then, using Sariel Har-Peled's algorithm, create an OBB around these points. This OBB is the over-approximation for this body, for this edge. 

% This overapproximation is then checked against an overapproximation of each obstacle. This overapproximation is an AABB of the obstacle. If the OBB and AABB do not intersect, then the edge is clearly free, in which case the edge is marked ''green''.  However, if there exists an intersection point, then we gain no information from the overapproximation, and mark the edge ''gray''. 

\subsection{Splines and spheres}

Sets of spheres have been widely used to approximate 3D shapes (e.g., \cite{sks-mssa-11}), though the perfect geometric fatness of spheres implies that a large set of spheres may be required to adequately approximate shapes with thin features. For this reason, we use constant size sets of spheres to under-approximate obstacles, but use splines extracted from the swept volumes of such spheres to approximate swept volumes.  

\begin{figure}[t!]
    \centering
    \includegraphics[width=\linewidth]{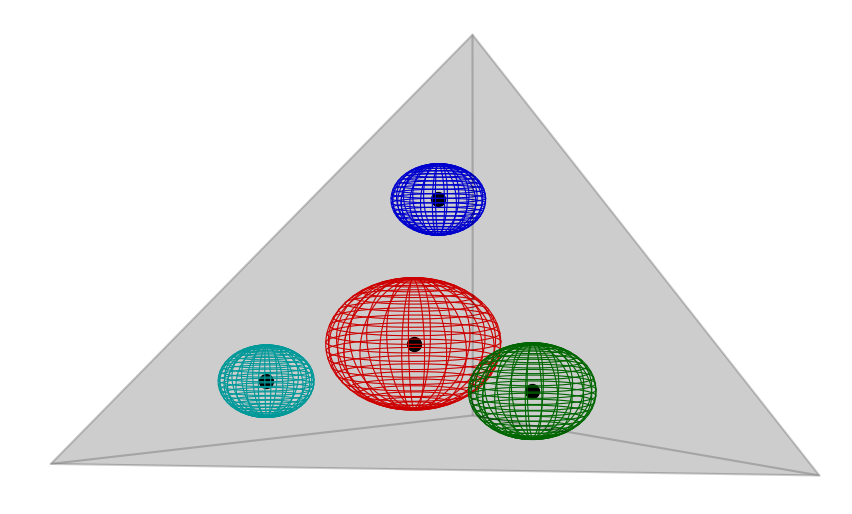}
    \caption{The output of Algorithm \ref{alg:sphere:generation} on an input tetrahedron $P$, $k=4$, and $j=4$.}
    \label{fig:sphere:example}
\end{figure}

\paragraph*{Sphere generation} In order to generate the spheres within a given body, we use a simple algorithm described in Algorithm \ref{alg:sphere:generation}. Since this simple algorithm is not one of our contributions, we do not describe its components in full. See Figure \ref{fig:sphere:example} for an example of four spheres generated in a tetrahedron using our code.  

\begin{algorithm}
\caption{Sphere generation}
\begin{algorithmic}[1]
    \State \textbf{Input:} Polyhedron $P$, number of samples $n$, number of spheres $k$, medial axis push steps $j$.
    \State \textbf{Output:} List of $k$ spheres contained in $P$
    \State $\texttt{spheres} \gets \{\}$
    \For{$i \gets 1$ to $n$}
        \State Generate a random point $p\in P$
        \For{$l \gets 1$ to $j$} \Comment{Pushing $p$ towards medial axis}
            \State $c \gets$ nearest point to $p$ in $\boundary{}P$ \Comment{\small Boundary of $P$}
            \State $opp \gets$ point opposite to $c$ on $\boundary{}P$
            \State $p \gets$ midway point between $c$ and $opp$
        \EndFor
        \State $r \gets$ distance between $p$ and the nearest point in $\boundary{}P$
        \State Append $\texttt{Sphere}(pt, r)$ to $\texttt{spheres}$
    \EndFor
    \State \textbf{return} LargeUnionK(\texttt{spheres}, k) 
\end{algorithmic}
\label{alg:sphere:generation}
\footnotesize
\textbf{Note:} \texttt{LargeUnionK($\texttt{spheres}, k$)} returns a set of $k$ spheres in $\texttt{spheres}$ such that the union of the area of these spheres is large. 
\end{algorithm}

\paragraph*{Spline generation}
After spheres have been generated for each of the bodies composing the robot, splines are constructed to under-approximate swept volumes.
Let $B$ be one of the polyhedral bodies composing the robot, and let $S=\Set{s_i = (c_i,r_i)}{1\leq i \leq k}$ be the set of spheres generated for $B$, with centers and radii $\set{c_i}_{i=1}^k$ $\set{r_i}_{i=1}^k$ respectively. 

First, we generate spline $S_{e,i}$ by concatenating the centers $c_i$ in every configuration of $\cfg(e)$. The complexity of $S_{e,i}$ depends on the length of $e$ and the resolution of the roadmap, even in cases where the $\FK(e)$ is relatively simple. We therefore use a curve simplification algorithm by Agarwal \etal \cite{ahmw-nltaa-05} to find ``shortcuts'' that maintain the under-approximation property.

Agarwal \etal{}'s algorithm uses a decision criterion (related to the desired approximation factor) when looking for shortcuts. To this end, we define \emph{good} and \emph{bad} segments. 
For two indices $i<j$ consider the line segment $l = c_{i}c_{j}$. $l$ is good if, for every index $m$, $i < m  < j$, the distance between point $c_m$ and $l$ is less than $r$, where $r$ is the radius of the sphere. 

\begin{figure}[t!]
    \centering
    \includegraphics[width=0.6\linewidth]{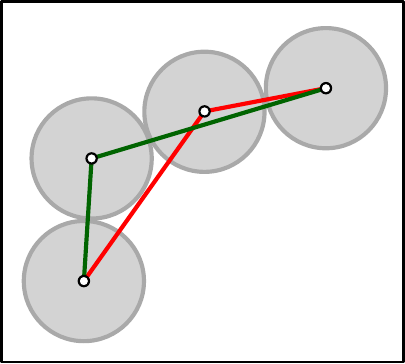}
    \caption{Illustration of the definitions used for Algorithm \ref{alg:shortcutting} for finding shortcuts in splines. One spline contains a bad segment (red), and the other only good segments (green).}
    \label{fig:goodbad}
\end{figure}

Formally, $l$ is good if 
\begin{equation*}
    \underset{i<m<j}{\max}\left(\dist(l,c_m)\right) < r
\end{equation*}
where $\dist(\cdot,\cdot)$ is the euclidean distance between two sets of points. $s$ is bad if it is not good. See Figure \ref{fig:goodbad} for an example. The pseudocode of the algorithm is given in Algorithm \ref{alg:shortcutting}.

\begin{algorithm}
\caption{Spline shortcutting}
\begin{algorithmic}
    \State \textbf{Input:} List of centers $\texttt{Centers}\subseteq \R^3$, radius $\texttt{r}\in \R^+$
    \State \textbf{Output:} List of shortcut points forming \texttt{spline}
    \State Initialize an empty list $\texttt{spline}$
    \State $n \gets$ size of \texttt{Centers}
    \State $i \gets 0$
    
    \While{$i < n$}
        \For{$j = 0$ to $\log_2(n - i) + 1$}
            \State $k \gets i + 2^j$
            \If{$\texttt{segment}[i, k]$ is \textit{bad}}
                \State $i \gets$ \texttt{GoodBadPt}($i,(i + 2^{(j-1)}), k$)  
                \State Append $i$ to \texttt{spline}
                \State break 
            \EndIf
        \EndFor
    \EndWhile

    \State \textbf{return} \texttt{spline}
\end{algorithmic}
\footnotesize
\textbf{Note:} \texttt{GoodBadPt($i, j, k$)} returns a point $l$ such that $j < l <k$ and segment $c_i c_{l}$ is good but $c_i c_{l+1}$ is bad. This is implemented using binary search.
\label{alg:shortcutting}
\end{algorithm}

% Figure of good and bad splines 

Each spline is checked against an inner-approximation of each obstacle. This inner-approximation uses a set of $k$ spheres that populate the obstacle. Since the spheres are entirely enclosed in the obstacle, if a sphere intersects a spline, that is a clear collision, making the edge invalid. Figure \ref{fig:rgg:example:mobile} shows obstacle $O_2$ being in collision with a spline, making the edge invalid, or ``red''.  Thus, the edge is marked ``red''. If there is no intersection, no information is gained from this inner-approximation and the edge is marked ``gray''. Figure \ref{fig:rgg:example:manip} shows obstacle $O_2$ is not in collision with the splines, but the edge is still invalid.  

If a robot has multiple bodies, each body is treated as its own robot and a set of splines is created for each body.

% A bad segment, is one where there exists at least one point, $x$, $i < x < j$, such that the distance between $s$ and $x$ is greater than $r$. 

% A figure where it shows good and bad segments would be nice here.

% \begin{table}[t!]
% %    \centering%
%     % \begin{center}
%     \input{exp_runtime_results}
%     % \end{center}
%     \caption{Preprocessing time (in seconds) for experiments described in Section \ref{sec:experiment}. Results were rounded down to the nearest integer}
%     \label{tab:results}
% \end{table}

\subsection{Implementation Details}

Our dynamic roadmap class that implements the $\RGG$ paradigm stores outer- and inner-approximations in two \AABB Trees that support intersection queries of \AABB{}s and spheres.
Inner-approximations created three spheres per robot body, and 6 for obstacles. The approximation parameter for computing \OBB{}s was $\delta = 0.1$.

The implementation of \DRM used is the implementation used in the experiments in \cite{almmmhpa-spitme-24}.
Since \DRM is parameterized by grid resolution we compared using a grid resolution $4$ (meaning cells of side length $4$). This resolution was chosen due to preprocessing time considerations. The preprocessing time required for finer resolution is somewhat excessive, and, since a resolution of 8 is too coarse to produce meaningful results, we used a finer resolution.

\paragraph*{Note on preprocessing time} We did not run the preprocessing phase enough times to justify reporting the results, but it is worth mentioning that the results replicate the findings in \cite{almmmhpa-spitme-24}, i.e., using swept volume approximations leads to a much faster preprocessing phase. Even for cells of side length 4 our method's preprocessing time was faster by a factor of $\sim10$. 

\section{Experiments}
\label{sec:experiment}

In this section we describe the experiments conducted to evaluate the usefulness of our implementation of the $\RGG$ approach. We compare the accuracy and runtime of our roadmap update method (\RGG) to those of \DRM, and use a brute force update function that uses \si{GAMMA}'s \si{PQP} algorithm \cite{lglm-fpqss-99} as a benchmark and an oracle for the ground truth of roadmap validities. 

\begin{figure}[t!]
    \centering
    \includegraphics[width=\linewidth]{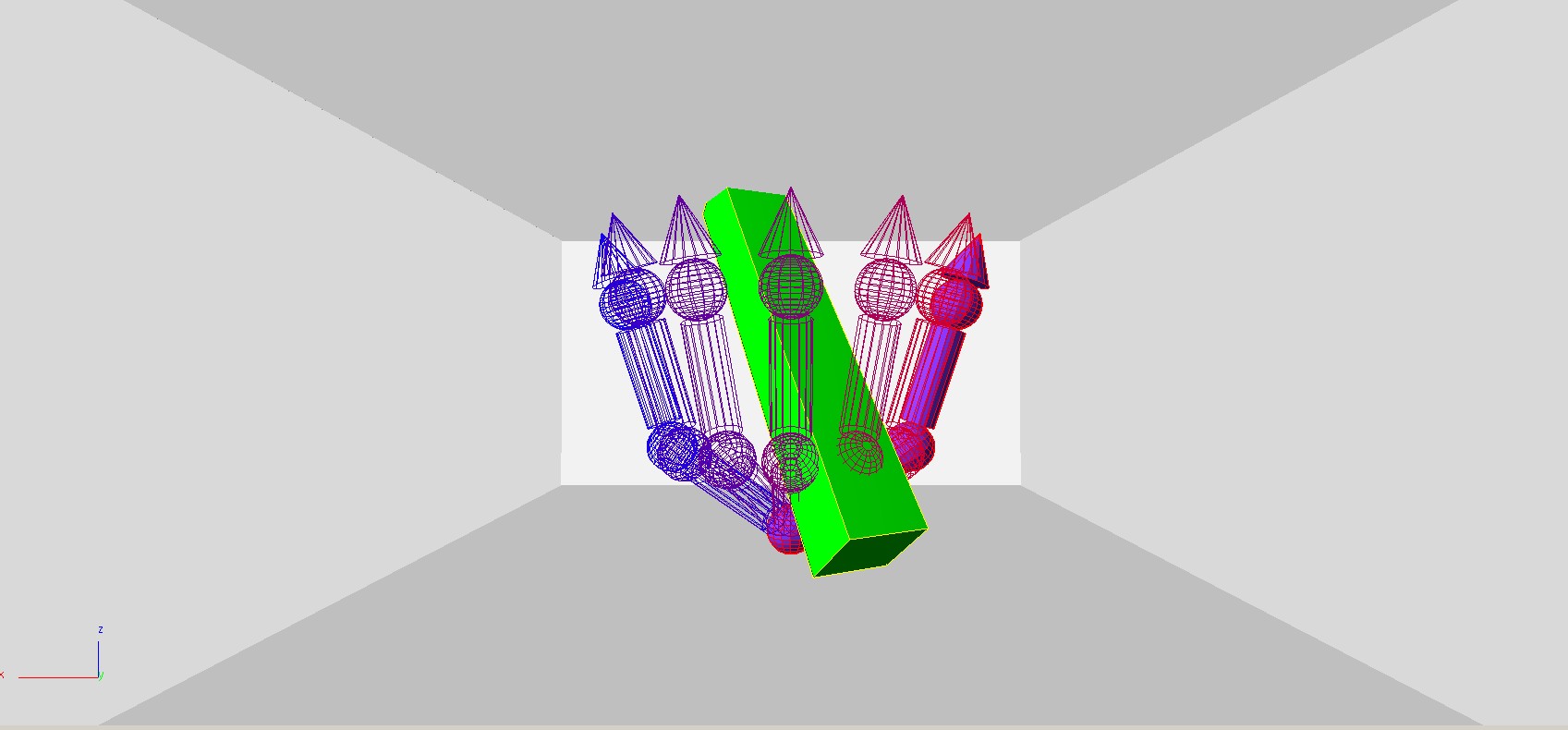}
    \caption{An illustration of an experiment described in Section \ref{sec:experiment}. Configurations of the simulated 6\DOF manipulator forming an edge in the workspace. The depicted edge is invalid due to the orientation of the rectangular prism obstacle.}
    \label{fig:experiment}
\end{figure}

The experiments simulate a fixed-base 6\DOF manipulator, composed of two links and an end-effector connected by three spherical joints, in a workspace composed of a bounding box of size $32\times32\times16$ containing a single rectangular prism obstacle. See Figure \ref{fig:experiment} for an illustration of the robot in the workspace.

In the preprocessing phase, a roadmap containing $\sim1000$ edges was generated using \PRM, and used to construct \RGG and \DRM dynamic roadmaps. This was followed by 100 repetitions of the following experiment. First, a random orientation was chosen for the obstacle, then, the \RGG and \DRM roadmaps update operations are called, and, lastly, every node and edge in the roadmap was validated by \si{PQP} in order to record the ground truth validities for that experiment.

\begin{figure*}[t!]
    \centering
    \includegraphics[width=\linewidth]{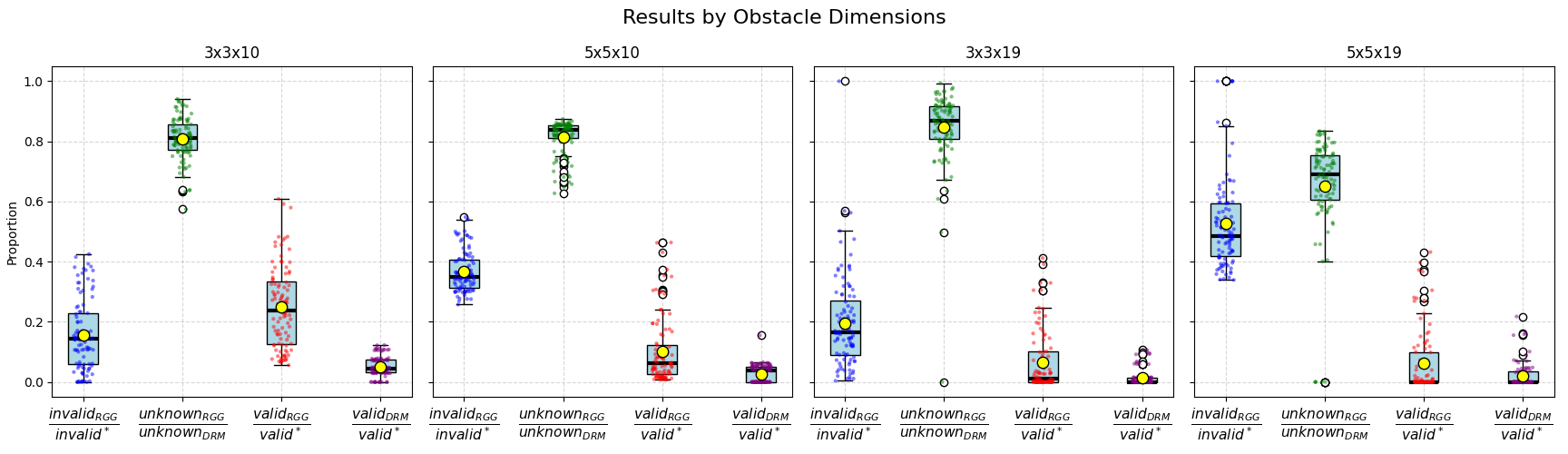}
    \caption{Box plots of the accuracy results from the experiments described in Section \ref{sec:experiment}. $valid^*$ and $invalid^*$ are, respectively, the ground truth numbers of valid and invalid edges computed by \si{PQP}, and $valid_{alg}$, $invalid_{alg}$, and $unknown_{alg}$ are, respectively, the numbers of valid, invalid and unknown edges labeled by algorithm $alg$. Each plot contains information for a different obstacle (see dimensions above plots), the mean is shown as a yellow dot, and the median as a black line segment. The different ratios, denoted below each bar, show how accurate \RGG and \DRM are when determining the validity of edges.}
    \label{fig:results:ratio}
\end{figure*}

\begin{figure}[t!]
    \centering
    \includegraphics[height=2.5in]{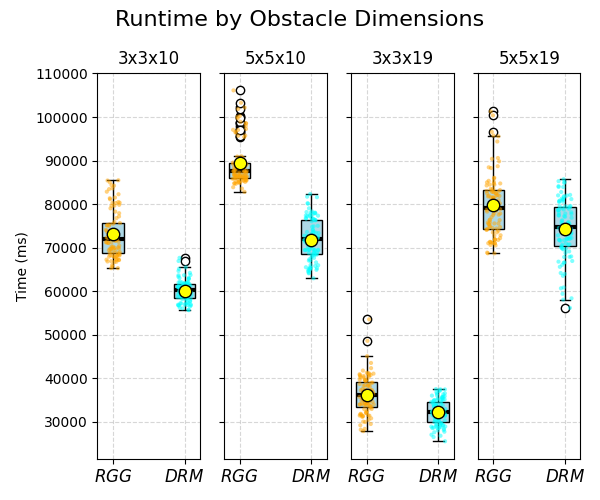}%
    \caption{Runtimes of \RGG and \DRM in the experiments described in Section \ref{sec:experiment}.}
    \label{fig:results:time}
\end{figure}

We report the fraction of valid edges labeled as valid by \RGG and \DRM out of the set of valid edges, the fraction of invalid edges labeled as invalid by \RGG out of the set of valid edges (as \DRM does not label any edge as invalid), and the runtime for each update operation, including the brute force application of \si{PQP}. We report the mean, median, $\min$, $\max$, and standard deviation for each of these results. Note that no edge was incorrectly labeled by either \RGG or \DRM. The information is graphically displayed in Figures \ref{fig:results:ratio} and \ref{fig:results:time}.

\paragraph*{Additional results}
The runtimes of the brute force \si{PQP} updates were higher than both methods by two orders of magnitude.
All data and exact values for the different descriptive statistics will be made available by the authors upon request. 

% \begin{figure}[t!]
%     \centering
%     \subfloat[Update time \RGG / \DRM]{%
%        \includegraphics[height=2.5in]%
%        {figs/time_by_obst_dim.png}%
%        \label{fig:results:time:rgg:drm}} 
%     \hfil
%     \subfloat[Brute force update time]%
%     {\includegraphics[height=2.5in]%
%        {figs/time_brute_force}%
%        \label{fig:results:time:bf}}
%     \caption{Time}
%     \label{fig:results:time}
% \end{figure}

\paragraph*{Experimental setup} All experiments were run on a desktop computer with an Intel Core i7-14700F processor. The Parasol Planning Library (PPL \cite{open-ppl}) implementations were used for all motion planning functions and algorithms.

\section{Discussion}
\label{sec:discussion}

    The accuracy results in Figure \ref{fig:results:ratio} show the accuracy advantage of \RGG over \DRM. For all four obstacles the mean, median, and max values of the ratio $\frac{valid_{\RGG}}{valid^*}$ are higher than $\frac{valid_{\DRM}}{valid^*}$, meaning that \RGG was able to label more edges as valid. As a reminder, all labels were tested against those of \si{PQP} to ensure correctness. The relatively low levels of detection of valid edges stem from the adversarial choice of obstacles. Both methods use \AABB{}s as outer-approximations of obstacles, and thus perform poorly in the presence of thin obstacles.

    The values of the ratio $\frac{invalid_{\RGG}}{invalid^*}$ indicate a marked advantage of \RGG over \DRM and SPITE, as \RGG correctly identifies a large fraction of invalid edges in many of the experiments.

    The values of the ratio $\frac{unknown_{\RGG}}{unknown_{\DRM}}$ show that, on average, \RGG correctly labels between 15-35\% more edges, reduces the amount of unknown edges after update operations, and never performs worse than \DRM.

    Finally, the runtime results, shown in Figure \ref{fig:results:time}, show a mild advantage for \DRM. This shows that much of the work done by \RGG is focused on finding invalid edges, as the experiments comparing SPITE, which is, excluding the difference between cigars and \OBB{}s, the outer-approximation component of \RGG, to \DRM suggest SPITE to be faster. 

\subsection{Future work}
Our goal is to adapt this work into a real-time multi-query motion planner such as SPITE, that would be able both to react to fast moving agents and obstacles, and use trajectory estimations to reduce uncertainty.

Although we have not mentioned this consideration in Section \ref{sec:method}, the choice of approximation shapes was also made with the prospect of serialization for SIMD and GPU utilization in mind. Since collision detection and proximity queries for \OBB{}s, splines, spheres, and \AABB{}s are all serializable, our method can be extended to a real-time system capable of handling highly complex and dynamic environments.

At the same time, we will include different approximation methods for both the obstacles and the robot. We will incorporate hierarchical decomposition schemes that will allow a tradeoff between update runtime and accuracy and improve performance on CPUs.

\bibliographystyle{plain}
\bibliography{robotics}
\end{document}